\pdfoutput=1

\documentclass[11pt]{article}

\usepackage{float}
\usepackage{amsmath}
\newcommand{\citebracket}[1]{[\citenum{#1}]}
\usepackage{booktabs}
\usepackage{enumitem} 
\usepackage{amsmath, amssymb}

\usepackage[final]{acl}

\usepackage{times}
\usepackage{latexsym}
\usepackage{hyperref}
\usepackage{tablefootnote}
\usepackage{enumitem}

\usepackage[T1]{fontenc}

\usepackage[utf8]{inputenc}

\usepackage{microtype}

\usepackage{inconsolata}

\usepackage{graphicx}

\title{SENTRA: Selected-Next-Token Transformer for LLM Text Detection}

\author{
  Mitchell Plyler \\
  Mozilla Corporation \\
  \texttt{mlplyler@mozilla.com} \And
  Yilun Zhang \\
  Mozilla Corporation \\
  \texttt{tzhang@mozilla.com} \And
  Alexander Tuzhilin \\
  New York University \\
  \texttt{at2@stern.nyu.edu} \AND
  Saoud Khalifah\dag \\
  Ciphero AI \\
  \texttt{saoud@ciphero.ai} \And
  Sen Tian\dag \\
  Ciphero AI \\
  \texttt{sen@ciphero.ai}
}

\begin{document}

\maketitle
\begin{abstract}
LLMs are becoming increasingly capable and widespread. Consequently, the potential and reality of their misuse is also growing. In this work, we address the problem of detecting LLM-generated text that is not explicitly declared as such. We present a novel, general-purpose, and supervised LLM text detector, \emph{SElected-Next-Token tRAnsformer (SENTRA)}. SENTRA is a Transformer-based encoder leveraging selected-next-token-probability sequences and utilizing contrastive pre-training on large amounts of unlabeled data. Our experiments on three popular public datasets across 24 domains of text demonstrate SENTRA is a general-purpose classifier that significantly outperforms popular baselines in the out-of-domain setting. 
\end{abstract}

\renewcommand{\thefootnote}{}
\footnotetext{\textbf{\dag} {Work performed at Mozilla Corporation.}}
\footnotetext{\textbf{Software:} {https://github.com/Firefox-AI/SENTRA/}}

\renewcommand{\thefootnote}{\arabic{footnote}}

\section{Introduction}

The problem of determining whether a text has been generated by an LLM or written by a human has been widely studied in both academia \cite{tang_science_2023} and industry. Several commercial-level automated text detection systems have been developed, including GPTZero \cite{tian2023gptzero}, Originality \cite{originalityai}, Sapling \cite{saplingAI}, and Reality Defender \cite{reality_defender}. Although significant progress has been made in detecting LLM-generated text over the past several years, these systems remain far from perfect and are often unreliable. A major limitation is their brittleness: they can perform well on certain types of LLM-generated text but fail catastrophically in other cases \cite{dugan_raid_2024}. This issue is particularly pronounced when operating in a real world scenario, where models must handle out-of-domain (OOD) data, different LLM generators, or various LLM "attacks" \cite{dugan_raid_2024, zhou2024humanizingmachinegeneratedcontentevading}. Therefore, it is crucial to develop more generalizable methods that deliver reliable performance across these settings.  

The probability of a document under an LLM's model can be measured by auto-regressively feeding the document's tokens into the LLM and observing the predicted probabilities for each token. This process produces a sequence of values called selected-next-token-probabilities (SNTP) that has been extensively used in prior work on LLM-generated text detection \cite{guo_how_2023,hans_spotting_2024, verma_ghostbuster_2023}. These prior works primarily rely on either heuristics (handcrafted functions) applied to SNTP sequences or linear models trained on expert-derived features \cite{hans_spotting_2024, verma_ghostbuster_2023}. In contrast, our proposed approach encodes SNTP sequences using a Transformer model pre-trained on unlabeled data, leveraging the expressivity of Transformers to directly learn a representation of the probability that a single or multiple LLMs assign to tokens in a document. More specifically, we propose \emph{SElected-Next-Token tRAnsformer (SENTRA)}, a method for detecting LLM-generated text that directly learns a detection function in a supervised manner from SNTP sequences.  
This method utilizes a novel Transformer-based architecture with a contrastive pre-training mechanism. The learned representation can be fine-tuned on labeled data to create a supervised model that distinguishes LLM-generated texts from human-written texts. 

For the LLM-text-detection task, supervised detectors have been shown to generalize poorly outside the training distribution \cite{dugan_raid_2024}.  Prior supervised methods typically leverage raw tokens as input and tend to overfit to token selections in a document. Heuristic or linear models on SNTP input have been shown to generalize well, but these simple models lack the expressivity to fully exploit the information in the SNTP sequences. Our SENTRA network addresses this issue by learning generalizable functions on SNTP. We show empirically that the supervised method presented in this paper generalizes to unseen domains better than both supervised and unsupervised baselines by leveraging our proposed Transformer-based architecture, thus demonstrating greater generalization to distribution shifts. 

In this paper, we demonstrate the following:
\begin{itemize}[noitemsep,leftmargin=*]
    \item Detectors utilizing SENTRA as their encoder \emph{generalize} well to domains outside of the training distribution(s).
    \item Contrastive pre-training of SENTRA leads to \emph{improved generalization} results on new domains.
    \item SENTRA outperforms all studied baselines in out-of-domain evaluations on three widely used benchmark datasets.
\end{itemize}
Because of the number of possible domains, improving out-of-domain generalization is the most important task to achieve LLM generated text detection in the wild.

\section{Related Work}
\label{related:work}
With the rise of LLMs, significant research has been conducted on accurately detecting text generated by these models \cite{tang_science_2023}.  At a high level, these detectors can be categorized into three approaches: watermarking, unsupervised (or zero-shot) detection, and supervised detection. Watermarking generally relies on the LLM deliberately embedding identifiable traces in its output \cite{liu_survey_2025}. In this work, we focus on the general detection problem, including cases involving non-cooperative LLMs; therefore, we do not consider watermarking as a point of comparison. Unsupervised methods typically leverage metrics computed by an LLM on the target document. These methods can be further divided into white-box detection, where the candidate LLM is known \cite{mitchell_detectgpt_2023}, and black-box detection, where the candidate LLM is unknown \cite{tang_science_2023}. Given our focus on the general detection problem, we prioritize black-box detection methods. Supervised methods, on the other hand, involve collecting a corpus of human-written and LLM-generated text samples, which are then used to train the detection models \cite{verma_ghostbuster_2023, soto_few-shot_2024}.

Selected-next-token-probabilities (SNTP) have been widely used for LLM detection in both white and black box settings \cite{guo_how_2023, hans_spotting_2024, verma_ghostbuster_2023}. Perplexity \cite{jelinek1977perplexity} is a commonly used metric to evaluate an LLM's ability to model a given dataset. In the context of AI detection, a lower perplexity score on a document indicates an LLM "fits" a document and this indicates a higher likelihood the document was LLM-generated. Conversely, a higher perplexity score suggests the LLM's probability model does not fit or accurately represent the candidate text, implying a lower likelihood that the text was generated by the LLM \cite{guo_how_2023}.

\begin{figure*}[h]
    \centering
    \includegraphics[width=\linewidth]{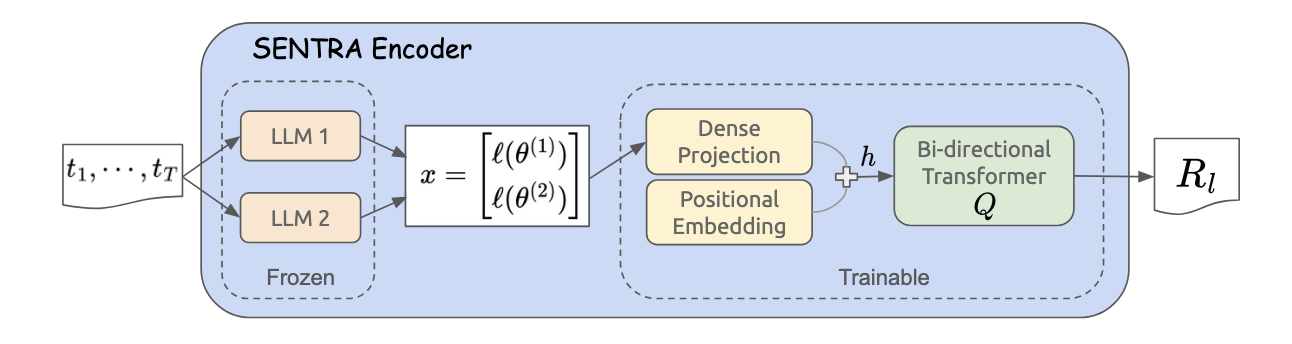}
    \caption{SENTRA leverages the selected-next-token-probabilities from two frozen LLMs. These two sequences of logits are concatenated into a vector. Each of these vectors are projected to the dimension of the bi-directional Transformer.
    }
    \label{fig:sntt}
\end{figure*}

Some detectors use multiple sequences of SNTP for the detection task \cite{verma_ghostbuster_2023, hans_spotting_2024}. \citet{verma_ghostbuster_2023} leveraged SNTPs from two Markov models, along with an LLM’s SNTP, extracted features, and a forward feature selection scheme as inputs to a linear classifier. In contrast to \citet{guo_how_2023}, \citet{hans_spotting_2024} argued that relying solely on the perplexity score for LLM-generated content detection can be misleading. Although human-authored text generally results in higher perplexity, prompts can significantly influence perplexity values. The authors highlighted the "capybara problem", where the absence of a prompt can cause an LLM-generated response to have higher perplexity, leading to false detections. They addressed this issue by introducing \emph{cross-perplexity} as a normalizing factor to calibrate for prompts that yield high perplexity. 

DetectGPT is an unsupervised method based on the idea that texts generated by LLMs tend to "occupy negative curvature regions of the model's log probability function" \cite{mitchell_detectgpt_2023}. The method generates perturbations of the sample text using a smaller model and compares the log probability of the sample text to that of the perturbations. Fast-DetectGPT replaces the perturbations in DetectGPT with a more efficient sampling step \cite{bao_fast-detectgpt_2024}.
\citet{nguyen-son_simllm_2024} observed that the similarity between a sample and its counterpart generation is notably higher than the similarity between the counterpart and another independent regeneration. They demonstrated that this difference in similarity is useful for detection. 

The most common supervised baseline for LLM-generated text detection is a RoBERTa classifier \cite{liu_roberta_2019} trained on a corpus of labeled text, where each document is marked as either human-written or LLM-generated. Several studies have expanded on this approach to supervised text-based classification. \citet{yu-etal-2024-text-flu} trained a feed-forward classifier with two hidden layers using intrinsic features derived from Transformer hidden states, determined via KL-divergence.  
\citet{tian_multiscale_2024} address the challenge of detecting short texts by treating short samples in the training corpus as partially "unlabeled". 
\citet{hu2023radar} employed adversarial learning to improve the robustness of their RoBERTa-based classifier against paraphrase attacks.

Several publications have explored contrastive training for the LLM detection task \cite{bhattacharjee_conda_2023, bhattacharjee_eagle_2024, soto_few-shot_2024, guo_detective_2024}. These studies use contrastive pre-training for a text Transformer, which is chosen to be RoBERTa \cite{liu_roberta_2019} in many cases, to guide the network toward a representation more useful for LLM-generated text detection. Furthermore, many prior contrastive training strategies focus on identifying stylometric features \cite{soto_few-shot_2024, guo_detective_2024}, while other studies extract stylometric features directly and train classifiers using those features \cite{kumarage_stylometric_2023}.  Rather than focusing on text representations, our method is mainly designed to produce useful SNTP representations and, thus, proposes a different contrastive pre-training scheme that compares textual representations with those of the SNTP Transformer.

However, SNTP and supervised methods have been shown, both intuitively and empirically, to struggle with generalization to unseen domains \cite{li_mage_2024, roussinov-etal-2025-controlling}. 

For instance, \citet{lai_adaptive_2024} applied adaptive ensemble algorithms to enhance model performance in OOD scenario. Meanwhile, \citet{guo_detective_2024} and \citet{soto_few-shot_2024}, recognizing the limited number of widely adopted general-purpose AI assistants, proposed to train an embedding model to learn the writing style of LLMs, and thereby improving the detection accuracy.

Prior work has shown SNTP to be an effective input for identifying LLM generated text \cite{guo_how_2023, hans_spotting_2024, verma_ghostbuster_2023}, but they rely on relatively simple metrics or heuristics. In this paper, we propose a Transformer-based SENTRA model that
learns a representation of SNTP sequences used for more effective training of detection models that better generalize to unseen domains.

\section{Methodology}

\begin{figure*}[h]
    \centering
    \includegraphics[width=\linewidth]{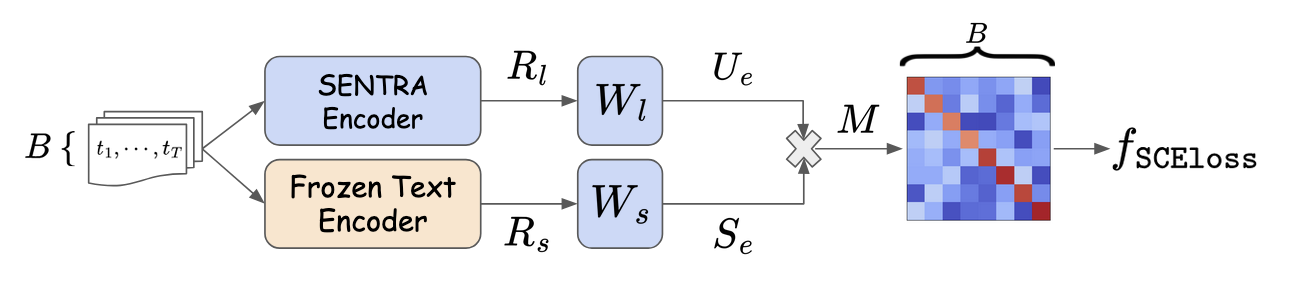}
    \caption{Pre-training: the outputs of SENTRA and a frozen text encoder go through linear layers, ($W_s$ and $W_l$) respectively, and normalization before a matrix multiplication (\texttt{matmul}) operation to produce the similarity matrix $M$. Blue and orange blocks indicate trainable and frozen components respectively.}
    \label{fig:pretraining}
\end{figure*}

\subsection{Overview of the SENTRA Method}

Consider a document $t$ consisting of an input sequence of $T$ tokens
$t = (t_1, t_2, \cdots, t_T)$.
Assuming an LLM has parameters $\theta$, the probability of document $t$ given this LLM can be specified as
\begin{equation}
    P(t_1, t_2, \cdots, t_T | \theta) = \prod_{t=1}^T q_i(\theta)
\end{equation}
where 
\begin{equation} \label{eq:q}
    q_i(\theta) = P(t_i \mid t_1, t_2, \cdots, t_{i-1}; \theta)
\end{equation}
is the probability of token $t_i$ given the preceding tokens $(t_1, t_2, \cdots, t_{i-1})$. We denote the observed sequence of selected-next-token-probabilities (SNTP) as 
\begin{equation}
q(\theta)=\left(q_1(\theta), q_2(\theta), \cdots, q_T(\theta)\right).
\end{equation}

It is common, and done in this work, to use the $\log$ representation of these sequences
\begin{equation}
    \ell_i(\theta) = -\log q_i(\theta)
\end{equation}
where $\ell$ is the $\log$ of the SNTP sequences.

Prior work, reviewed in Section~\ref{related:work}, has proposed various heuristic functions on these sequences that are useful in detecting LLM-generated text \cite{guo_how_2023,hans_spotting_2024}. 
SENTRA replaces these heuristic functions on SNTP sequence(s) with a neural network, as shown in Figure \ref{fig:sntt} illustrating our proposed method. In particular, we leverage $k$ LLMs, each with parameters $\theta^{(k)}$ to produce SNTP sequences $\ell^{(k)}$ and for a candidate document with $T$ tokens using process in Equation \ref{eq:q}. The $k$ sequences are concatenated to form input sequence $x$.
Note that in Figure \ref{fig:sntt}, $k=2$. In this work, we focus on the $k=2$ case. Setting $k>1$ allows the model to learn from similarities and deviations in SNTP sequences produced by LLMs. This comparison was a key idea in \cite{li_spotting_2024}, and following that work, we focus on the $k=2$ case where the two LLMs share a tokenizer. This allows the SNTP sequences to be aligned.

Instead of token embeddings often seen in Transformer architectures \cite{devlin-etal-2019-bert}, 
each token-indexed representation $x_t \in x$ is independently projected using a fully connected layer. 
\begin{equation}
    h_t = f(Wx_t + b) + Z_t
\end{equation}
where $h$ is the dense embedding representation, $f$ is the ReLU activation function, $W$ is the weight matrix, $b$ is the bias, and $Z_t$ are $Z \in \mathbb{R}^{T\times D}$ learned positional embeddings.
This transformation results in a representation of size $T \times D$ for a single document. Note a learned [CLS] representation $h_{[CLS]} \in \mathbb{R}^D$ is pre-pended to the sequence before the positional embeddings are applied. This representation $h_t$ is passed through a bi-directional Transformer \cite{devlin-etal-2019-bert} $Q$, as shown in Figure~\ref{fig:sntt}.

The output of SENTRA is a learned representation over SNTP, capturing the probability assigned by two LLMs to the tokens in a document. For classification, we use the representation at the \texttt{[CLS]} token and append a classification head. This Transformer produces our SENTRA representation $R_l$ over SNTP sequences.
\begin{equation}
    R_l = Q(h)
\end{equation}

where $R_l$ is a $D$ dimensional representation of the document over the token length $T$.

In summary, SENTRA is the first Transformer-based encoder to systematically learn a useful representation of SNTP sequences. Similar to many Transformer-based approaches \cite{devlin-etal-2019-bert, clip}, that have traditionally used different modalities of input information, we demonstrate in Section \ref{sec:contrastive} that our method can leverage large quantities of unlabeled data to enhance this learned representation.

\subsection{SENTRA Contrastive Pre-Training} \label{sec:contrastive}

We further explore the pre-training of SENTRA using unlabeled text data and demonstrate in Section \ref{sec:results} that it significantly improves SENTRA's performance. Notably, this pre-training scheme is reminiscent of CLIP \cite{clip}.
Figure \ref{fig:pretraining} illustrates our concept for pre-training SENTRA. We leverage off-the-shelf, pre-trained text representations to help SENTRA learn a useful representation of SNTP sequences. A document is encoded using both a pre-trained text encoder \cite{devlin-etal-2019-bert,liu_roberta_2019} and our SENTRA network, producing representations $R_l$ and $R_s$. These representations are projected to a joint embedding space, $U_e$ and $S_e$, using fully connected layers $C_l$ and $C_s$ for the text and SNTP representations respectively. 

\begin{equation}
\begin{aligned} 
U_e &= C_l(R_l) \\
S_e &= C_s(R_s)
\end{aligned} 
\end{equation}

After applying L2 normalization to $U_e$ and $S_e$ to control for scaling, we then compute a comparison matrix $M$
\begin{equation}
    M = (U_eS_e^T) e^r
\end{equation}
where $r$ is learned temperature scalar.

The two encoders learn to match representations of the same document within a batch $B$. Employing the contrastive learning objective, 
\begin{flalign}
\mathcal{L} = \frac{\mathcal{L}_s + \mathcal{L}_l}{2} &&
\end{flalign}

\begin{flalign}
\mathcal{L}_l = -\frac{1}{n} \sum_{i=1}^n \log \left( \frac{\exp(M_{ii})}{\sum_{j=1}^n \exp(M_{ij})} \right) &&
\end{flalign}

\begin{flalign}
\mathcal{L}_s = -\frac{1}{n} \sum_{j=1}^n \log \left( \frac{\exp(M_{jj})}{\sum_{i=1}^n \exp(M_{ij})} \right) &&
\end{flalign}
we then minimize the cross-entropy loss over the columns (text-to-SNTP), and rows (SNTP-to-text) of the comparison matrix $M$, using the ground truth text-SNTP pairings in the batch, $y=0,1,...B-1$.

The pre-training scheme effectively enables SENTRA to produce representations that align with those generated by the frozen text encoder, thereby yielding more useful representations of the $\ell^{k=1}$ and $\ell^{k=2}$ sequences. 

 In \cite{clip}'s work, the authors jointly trained text and image encoders from scratch. Unlike CLIP, which deals with text and images, we focus solely on text and on pre-training only the SENTRA SNTP encoder. To do this, we freeze a pre-trained text encoder and train only SENTRA and the contrastive embedding projections.

\subsection{Implementation}

\label{sec:implementation}
We implement our SENTRA model with eight attention heads, eight layers, and a hidden dimension of $768$ for a total of 57M parameters. The Transformer architecture and positional embeddings follow the same definitions as in BERT \cite{devlin-etal-2019-bert}. Before pre-training, the SENTRA parameters are randomly initialized. The frozen text encoder used for contrastive pre-training is initialized from RoBERTa \cite{liu_roberta_2019}. SENTRA is pre-trained on a 600K sample of Common Crawl data from RedPajama \cite{weber2024redpajama}. Pre-training is conducted for $20$ epochs with a batch size of $256$ and a maximum token length of $64$. We then continue contrastive training for $10$ epochs with a batch size of $128$ and a maximum token length of $512$ to pre-train the later position embeddings. The peak learning rate was set to $1\mathrm{e}-4$ for both phases. We use the AdamW \cite{loshchilov2018decoupled} optimizer with a weight decay of $1\mathrm{e}-2$ and set the contrastive learning temperature to $0.007$ \cite{chen2020simclr}. During fine-tuning, we initialize SENTRA from the pre-trained model, use a learning rate of $1\mathrm{e}-4$, a weight decay of $1\mathrm{e}-2$, and apply early stopping with a patience of two epochs on a validation dataset. 

As shown in Figure \ref{fig:sntt}, we implemented SENTRA with two SNTP sequences and therefore $k=2$. Following Binoculars \cite{hans_spotting_2024}, we use Falcon-7B 
and Falcon-7B-Instruct 
\cite{almazrouei_falcon_2023} to produce these sequences. We used a sequence of two SNTP because Binoculars showed it is useful for the detector to compare both SNTP, and we used the Falcon models specifically because Binoculars showed they worked well \cite{hans_spotting_2024}. 
During SENTRA training, the SNTP sequences are precomputed and cached. At inference, the computational complexity is dominated by the Falcon models. Because we use the same LLMs as Binoculars \cite{hans_spotting_2024} and our SENTRA encoder is small, our method has the same order of complexity as Binoculars. See Appendix \ref{app:complexity} for additional details.

\section{Experiments}

\subsection{Datasets} \label{sec:datasets}

If we define text similar to the training data distribution as in-domain and text that is dissimilar as out-of-domain, it is well established supervised LLM detection methods perform significantly better in-domain than out-of-domain \cite{dugan_raid_2024}. However, a model designed for LLM-generated text detection in real world scenarios will inevitably encounter out-of-domain texts. For this reason, this work focuses on \emph{out-of-domain experiments}, where key subsets of data are withheld from the training dataset.

To evaluate the effectiveness of our proposed method, we used three publicly available datasets: RAID \cite{dugan_raid_2024}, M4GT \cite{wang_m4gt-bench_2024} and MAGE \cite{li_mage_2024}, focusing exclusively on English-language data.

\textbf{RAID:} The full RAID dataset contains over 6 million human- and LLM-generated texts spanning $8$ domains, $11$ LLM models, multiple decoding strategies, penalties, and $11$ adversarial attack types. We down-sampled it to $500$K instances before performing out-of-domain split sampling. With the included attacks, the RAID dataset also assesses the effectiveness of different supervised baseline methods against adversarial attacks under the in-attack setup.

\textbf{M4GT:} An extension of M4 \cite{wang_m4_2024}, the M4GT dataset is a multi-domain and multi-LLM-generator corpus comprising data from $6$ domains, $9$ LLMs, and $3$ different detection tasks.

\textbf{MAGE:} The MAGE dataset covers $10$ content domains, with data generated by $27$ LLMs using $3$ different prompting strategies. It is specifically designed to assess out-of-distribution generalization capability. We use the "Unseen Domains" evaluation from \cite{li_mage_2024}.

Each dataset is further split into training, validation and test sets. For MAGE, we used the published split. To mitigate the label imbalance problem, the train and validation splits are balance-sampled to ensure an equal number of human- and LLM-generated texts. This was achieved by down-sampling the majority class to match the size of the minority class within split. Addressing this imbalance is crucial for two reasons: 1) the percentage of LLM-generated text is over $97\%$ in the RAID dataset by design; 2) across the three datasets, the proportion of LLM-generated text varies significantly. The average train and validation set sizes show how much data went into the training of the supervised methods while ensuring class balance, providing a clear comparison to the total dataset size. The MAGE dataset has significantly shorter texts and this adds difficulties to the detection task \cite{tian_multiscale_2024, fraser2024detectingaigeneratedtextfactors}.

Beyond out-of-domain evalution, we further assessed our method in an out-of-LLM (OOLLM) setup using MAGE's out-of-LLM testbed which contains 7 LLM splits. Table~\ref{tbl:dataset} contains detailed statistics on the evaluation datasets. For fair comparison across methods, we use the first $512$ tokens from each document in all datasets.

\subsection{Baseline Methods}
\label{baseline}

We evaluated and compared the performance of our approach against multiple existing methods, including zero-shot, embedding-based, and supervised detectors. For zero-shot, we selected \textbf{perplexity} \cite{guo_how_2023}, \textbf{Fast-DetectGPT} \cite{bao_fast-detectgpt_2024}, and \textbf{Binoculars} \cite{hans_spotting_2024} detectors. For embedding-based detectors, we selected \textbf{UAR} \cite{soto_few-shot_2024} and evaluated both its Multi-LLM and Multi-domain models. For supervised detectors, we chose \textbf{RoBERTa} \cite{liu_roberta_2019} with direct fine-tuning, \textbf{Ghostbuster} \cite{verma_ghostbuster_2023} which trains a logistic regression classifier on forward-selected crafted log-probability features, and \textbf{Text Fluoroscopy} \cite{yu-etal-2024-text-flu} which utilizes intrinsic features. For RoBERTa, we used the same settings as the fine-tuning of SENTRA: a learning rate of $1\mathrm{e}-4$, a weight decay of $1\mathrm{e}-2$, and a patience of two epochs.

We used Falcon-7B and Falcon-7B-Instruct across all baseline methods that required LLMs, except for Fast-DetectGPT where we followed its black-box setting. Appendix \ref{sec:appendixA} provides a detailed description of the setup, assumptions and modifications made for each baseline method.

We compared the baseline methods mentioned above with our proposed methods. We present results from two SENTRA encoder variations, R-SENTRA and SENTRA. R-SENTRA has all non-LLM weights in SENTRA encoder initialized at random (without pre-training), whereas the full SENTRA model has those weights pre-trained as described in Section \ref{sec:implementation}.

\begin{table*}[h]
\centering
\begin{tabular}{lcccccccc}
& \multicolumn{2}{c}{RAID-OOD} & \multicolumn{2}{c}{M4GT-OOD} & \multicolumn{2}{c}{MAGE-OOD} & \multicolumn{2}{c}{MAGE-OOLLM} \\
\cmidrule(lr){2-3} \cmidrule(lr){4-5} \cmidrule(lr){6-7} \cmidrule(lr){8-9}
& Avg & W & Avg & W & Avg & W & Avg & W \\
\midrule
r-SENTRA &90.9 & 85.5 & 92.8 & 83.9 & 93.8 & 84.6 & 93.5 & \textbf{89.9}\\ 
SENTRA &\textbf{92.5} & \textbf{87.0} & \textbf{93.0} & \textbf{87.1} & \textbf{94.2} & \textbf{86.0} & \textbf{93.6} & 88.0\\ 
\bottomrule
\end{tabular}
\caption{Effect of Pre-training on SENTRA performance. Results are the average (Avg) and worst (W) AUROC across the domains in the evaluation.}
\label{tbl:ablation_pre}
\end{table*}

Interestingly, prompting an LLM to do the LLM-text detection task is not well studied and does not appear in standard benchmarking work \cite{dugan_raid_2024, wang_m4_2024, li_mage_2024}. We performed a small case study to evaluate how a SOTA LLM, GPT-4o \cite{openai_gpt-4o_2024}, and a reasoning model, o1 \cite{openai_openai_2024}, could perform on a sample of the OOD datasets. We were unable, due to the high cost of these APIs, to run the full evaluation datasets through these models and therefore chose to randomly sample from the full datasets and perform a fair comparison on the smaller test sets. The evaluation results for the GPT4-o and o1 LLMs and their comparison with SENTRA performance are reported in Section \ref{sec:prompting}.

\subsection{Ablation Study}

\begin{table}[h]
    \centering
    {
    \begin{tabular}{lcc} 
& Avg & W \\ 
\hline
r-SENTRA &\textbf{92.8} & \textbf{83.9} \\ 
\midrule
\hspace{1em} $-$ Base LLM &89.4 & 81.8 \\ 
\hspace{1em} $-$ Instruct LLM &88.1 & 74.1 \\ 
\midrule
\hspace{1em} $-$ Falcon $+$ Qwen-2.5-3b &89.3 & 75.0 \\ 
\hspace{1em} $-$ Falcon $+$ Gemma-3-1b &91.2 & 82.7 \\ 
\bottomrule
\end{tabular}
    } 
    \caption{Ablation Study. Results show the average (Avg) and worst (W) domain AUROC on the M4GT dataset.  The top section, r-SENTRA, is our method without pre-training. The second section shows the effect of dropping each of the two frozen LLMs. The last section shows the effect of swapping the Falcon-7b models for different pairs of LLMs. }
    \label{tbl:ablation}
\end{table}

Table~\ref{tbl:ablation_pre} shows the effect of pre-training SENTRA on all datasets. r-SENTRA is the "raw" SENTRA showing the architecture's performance without pre-training on any dataset and then evaluating on the M4GT dataset. Across the four datasets, the average and worst-case performance over the domains was increased after pre-training. This shows the contrastive pre-training method presented in Figure \ref{fig:pretraining} is an effective method for improving SENTRA as an encoder for the LLM text detection.

Table~\ref{tbl:ablation} presents an ablation study on SENTRA components. Rows 2 and 3 of Table~\ref{tbl:ablation} show the AUROC performance metric after removing each of the two LLMs used to create SENTRA's SNTP input (see Figure~\ref{fig:sntt}).  Rows 4 and 5 of the table show the results when the Falcon-7b models \cite{almazrouei_falcon_2023} are replaced by different pairs of LLMs: Qwen-2.5-3b \cite{qwen_qwen25_2025} and Gemma3-1b \cite{team_gemma_2025}. From the results, we can see that Gemma3-1b \cite{team_gemma_2025} is competitive with Falcon-7b, and could be an alternative for more compute constrained environments. These choices in LLMs are by no means an exhaustive search, and this ablation shows SENTRA can work with other LLM pairs while echoing Binocular's result showing Falcon-7b is particularly effective \cite{hans_spotting_2024}.

\subsection{Results} 
\label{sec:results}

\begin{table*}[ht]
\centering
\begin{tabular}{lcccccccc}
& \multicolumn{2}{c}{RAID-OOD} & \multicolumn{2}{c}{M4GT-OOD} & \multicolumn{2}{c}{MAGE-OOD} & \multicolumn{2}{c}{MAGE-OOLLM} \\
\cmidrule(lr){2-3} \cmidrule(lr){4-5} \cmidrule(lr){6-7} \cmidrule(lr){8-9}
& Avg & W & Avg & W & Avg & W & Avg & W \\
\midrule
RoBERTa \citebracket{liu_roberta_2019} &90.9 & 84.4 & 88.2 & 82.8 & 88.3 & 74.4 & 87.1 & 69.9\\ 
Text-Fluoroscopy \citebracket{yu-etal-2024-text-flu} &76.4 & 70.6 & 83.2 & 78.1 & 63.9 & 47.8 & 41.5 & 28.3\\ 
UAR-D \citebracket{soto_few-shot_2024}&81.7 & 71.4 & 75.3 & 63.9 & 63.4 & 40.5 & 71.7 & 65.8\\ 
UAR-L \citebracket{soto_few-shot_2024} &87.3 & 76.3 & 84.7 & 71.0 & 76.4 & 61.2 & 80.4 & 70.7\\ 
Ghostbuster \citebracket{verma_ghostbuster_2023} &84.7 & 74.1 & 87.8 & 73.3 & 79.2 & 65.0 & 68.5 & 34.3\\ 
\midrule
PPL \citebracket{guo_how_2023} &72.9 & 69.4 & 87.0 & 81.7 & 57.2 & 45.7 & 59.0 & 25.4\\ 
Binoculars \citebracket{hans_spotting_2024} &82.0 & 79.4 & 89.1 & 79.0 & 61.7 & 52.9 & 61.8 & 14.7\\   
Fast-DetectGPT \citebracket{bao_fast-detectgpt_2024}&78.6 & 75.6 & 87.4 & 79.1 & 63.0 & 54.9 & 37.9 & 2.8\\ 
\midrule
SENTRA & \textbf{92.5} & \textbf{87.0} & \textbf{93.0} & \textbf{87.1} & \textbf{94.2} & \textbf{86.0} & \textbf{93.6} & \textbf{88.0}\\ 
\bottomrule
\end{tabular}
\caption{Average (Avg) and worst (W) out-of-domain AUROC across the domains or LLMs. Methods in the top section are supervised while the methods in the second section are unsupervised. SENTRA is our method with pre-training. Results for non-deterministic methods are averaged over three random seeds. }
\label{tbl:aggregate}
\end{table*}

We measure performance of all the methods described in Section~\ref{baseline} on three out-of-domain and one out-of-LLM evaluation, and the average and worst-case AUROC results are presented in Table~\ref{tbl:aggregate}. For the supervised methods, these evaluations assess how well the LLM text detectors perform in real world scenarios, where data distributions differ from the training distribution. Detectors that remain more invariant across these evaluations are considered more robust to changes and variations in data, thus showing better generalization to unseen domains and generators.

Methods that are not zero-shot or linear models are inherently more stochastic; therefore, the UAR, RoBERTa, and SENTRA methods were ran over three random seeds. The main results in Table~\ref{tbl:aggregate} show the mean over these seeds. Mean and standard deviation over the seeds across all domains and evaluations are shown in Appendix \ref{app:additional_results}. On each evaluation, our performance metric is the mean or minimum over the domains. For each method, this requires training a separate model for each random seed, each domain, and each evaluation. Because of the combinations of methods, seeds, domains, and datasets, each additional run becomes very expensive, and therefore, we were limited to three runs on each evaluation. 

Table~\ref{tbl:aggregate} presents performance of different baselines measured by AUROC across different OOD test data for the RAID, M4GT and MAGE datasets (columns RAID-OOD, M4GT-OOD and MAGE-OOD in Table~\ref{tbl:aggregate} respectively) and for the OOLLM test data for the MAGE dataset (column MAGE-OOLLM in the table).
The top section of Table~\ref{tbl:aggregate} shows the performance of label-dependent methods while the second section shows the performance of heuristic methods.

Table~\ref{tbl:aggregate} shows that SENTRA outperformed all the baselines on average and in the worst case across the three OOD and one OOLLM evaluations. SENTRA achieved average AUROC performance improvements of 1.8\%, 5.4\% and 6.7\% for RAID \cite{dugan_raid_2024}, M4GT \cite{wang_m4gt-bench_2024} and MAGE \cite{li_mage_2024} out-of-domain datasets respectively, as compared to the second-best performing baseline. For the OOLLM evaluation, SENTRA showed a 7.5\% increase over the next best baseline. These results show SENTRA serves as a generalizable encoder for LLM detection models when one considers likely OOD or OOLLM distribution shifts. These results show, in the likely event your detector encounters a domain outside the training distribution, we expect SENTRA to have the best expected performance and best worst-case performance on those unseen domains.

Since LLMs became increasingly available and their usage has surged, interest in detection tools, such as those presented in this paper, has grown \cite{wu_survey_2023}. At the same time, countermeasures have emerged to attack these LLM text detectors, typically by altering LLM-generated text to elicit false negatives \cite{koike_outfox_2024}. \citet{dugan_raid_2024} demonstrated many attacks can significantly degrade detector performance. In that study, the best open-source tool, Binoculars \cite{hans_spotting_2024}, exhibited much stronger performance on non-attacked data than on attacked data. For the unsupervised methods, \cite{guo_how_2023,hans_spotting_2024, bao_fast-detectgpt_2024}, it is not immediately clear how to adapt the approach to a known attack. In contrast, for the supervised methods, the adaptation strategy is straightforward: train on attacked data. A model that is robust to a known attack, like the common paraphrase attack, should be able to detect LLM generated text even if that attack appears in a new domain. The RAID-OOD \cite{dugan_raid_2024} dataset demonstrates this situation where 11 attacks appear in the training and test sets. The results in Table~\ref{tbl:aggregate} show SENTRA outperformed other methods when training and evaluating in the out-of-domain scenario where known attacks are included.

\subsection{LLM Prompting Case Study}
\label{sec:prompting}

As part of our benchmarking, we also evaluated OpenAI's proprietary models, namely \texttt{gpt-4o-2024-08-06} ("4o") and \texttt{o1-2024-12-17} ("o1"), by prompting them directly to classify whether a given text was written by a human or generated by an AI. The prompt is included in Appendex \ref{app:case_study}.

To control inference costs, we limited the evaluation to 100 samples per domain/model, using the same datasets from the OOD and OOLLM experiments.
The evaluation results are presented in Table~\ref{tbl:openai_result}, alongside SENTRA’s performance. Overall, the reasoning-based o1 model demonstrated stronger detection capabilities than the standard 4o model, particularly on RAID-OOD and M4GT-OOD. Nevertheless, SENTRA consistently outperformed both OpenAI models across all datasets.

\begin{table*}[h]
    \centering
    \resizebox{\columnwidth}{!}{%
    \begin{tabular}{l|c|c|c}
        Dataset & 4o & o1 & SENTRA \\
        \hline
        RAID-OOD & 79.5 & 90.0 & \textbf{91.1} \\
        M4GT-OOD & 65.4 & 91.1 & \textbf{92.9} \\
        MAGE-OOD & 75.1 & 78.4 & \textbf{92.9} \\
        MAGE-OOLLM & 72.1 & 75.3 &\textbf{93.8} \\
    \end{tabular}
    }
    \caption{AUROC scores for OpenAI models and SENTRA. Best score per dataset is bolded.}
    \label{tbl:openai_result}
\end{table*}

This case study underscores the need for full and rigorous evaluation when assessing LLM performance on the task of AI-text detection.

\section{Conclusions}

In this paper, we proposed a novel general purpose supervised LLM text detector method SENTRA that is a Transformer-based encoder leveraging SNTP sequences and utilizing contrastive pre-training on large amounts of unlabeled data. We show this supervised method acting on SNTP input outperforms previously considered heuristic functions and other methods that rely on text input.
Since supervised detectors tend to perform better on data that is similar to their training distributions \cite{dugan_raid_2024}, it is essential to include a wide variety of domains when testing such general-purpose detectors. Therefore, we tested the performance of SENTRA on three public datasets RAID, M4GT and MAGE containing a broad range of different domains (24 in total) across various experimental settings and compared its performance with eight popular baselines. We also evaluated SENTRA and the baselines on a out-of-LLM evaluation.

We empirically demonstrated that SENTRA significantly outperformed all baselines in our studied experimental settings. On our three evaluation datasets, SENTRA outperformed all eight popular baselines for the average and the worst-case OOD scenarios.

These results show that SENTRA is a strong method for training LLM text detectors that can generalize well to unseen domains and LLM generators. Our ablation study showed performance of SENTRA increases when two frozen LLMs are used instead of one frozen LLM. We also demonstrated our contrastive pre-training strategy increased the performance of SENTRA on all out-of-domain evaluations. 

Because SENTRA is better able to handle these critical out-of-domain and out-of-LLM settings, these results demonstrate SENTRA is a general-purpose encoder that can serve as a foundation for the LLM text detector models.

\section{Limitations}
\label{sec:limitation}
In this work, we studied the effects of domain shifts on detection models. While these have significant impacts on detector performance, other factors can also influence results. Notably, prompt variation can have a large effect on detectors \cite{kumarage-etal-2023-reliable}. Many LLM detection benchmark datasets use prompt templates \cite{dugan_raid_2024} to generate their samples. However, these templates exhibit significantly less prompt variety than what a real-world detector is likely to encounter. Benchmark datasets with a broader range of prompting strategies are needed to further assess the robustness of detection methods.

We pre-trained our model on a relatively small sample of Common Crawl data. The volume of data and the amount of compute used for pre-training were small relative to what is typically used for foundation models \cite{liu_roberta_2019, clip}. It is very likely SENTRA could be significantly improved with additional pre-training on larger datasets.

\section{Ethical Considerations}
In this study, we did not observe any detector achieving perfect performance on any slice of data. Therefore, any detector will inherently make trade-offs between false positives and false negatives when deployed in real-world scenarios, such as plagiarism detection. Users of LLM detection technology should be aware that these detectors are not perfect.

\textbf{LLM Acknowledgement:}
We used ChatGPT for generating first iterations of some software snippets. We also consulted ChatGPT on the phrasing of some points in the paper and for catching some grammatical errors.

\bibliography{latex/SNTT.bib}

\appendix
\section{LLM Case Study Details} \label{app:case_study}
 At the time of writing, we estimated that evaluating the full datasets would cost approximately \$10,000 for GPT-4o and \$60,000 for o1 - several orders of magnitude more expensive than any other method considered. We therefore elected to sample the datasets and move them to a seperate study than the other methods.

We used the following system prompt to obtain both a label and a confidence score: \textit{"You are an expert in identifying whether text was written by a human or generated by an AI language model. You are tasked to identify if a provided text is written by a human or generated by an AI language model. Return your answer on the first line as one word only: 'Human' or 'AI'. On the second line, provide a confidence score between 0 and 1. Do not output anything else."}. The returned confidence score was interpreted as the model's probability of the predicted class. To compute AUROC fairly, scores were flipped for predictions labeled as "Human". Due to the stochastic nature of its reasoning mechanism, we ran o1 three times and averaged the results. For 4o, we set temperature = 0 to reduce randomness.

 We emphasize that prompt engineering was not a focus of this work; we did not explore alternative prompting strategies such as few-shot examples, chain-of-thought reasoning, or tailored instruction tuning. These results should therefore be viewed as a simple baseline reference rather than a comprehensive exploration of prompt-based detection. A more thorough investigation—including experiments on full datasets, alternative prompting methods, and other comprehensive settings—is left for future work.

\section{Additional Results and Experimental Notes} \label{app:additional_results}
Here we present mean and standard deviation across the three random seeds. We first show tables with AUROC as the metric. The later tables show class-weighted F1 score. When computing F1, we set the class threshold at $0.50$. Because unsupervised methods require tuning a classification threshold, we only include the supervised methods for the F1 score. Notice the threshold of $0.50$ is arbitrary. In practical settings, we have found threshold tuning to be a challenging and critical problem, but we found it to be mostly separate from evaluating the overall quality of a classifier. When deploying AI detection models in the wild, we found it useful to tune the threshold to a desired false positive rate on common crawl data before the release of GPT2. For these reasons, the main text of the paper focuses on a threshold agnostic metric: AUROC.

\begin{table*}[h]
    \centering
    \resizebox{\textwidth}{!}{%
    \begin{tabular}{lcccccrrrrr}
        \hline
        Dataset & Size & Domains & LLMs & Attks & A.Tokens & \% LLM-Gen & A.Train & A.Val & A.Test\\
        \hline
        RAID-OOD & 500,000   & 8  & 11     & 11 & 712 & 97.16\%  & 22,398  & 2,488  & 62,500 \\
        M4GT-OOD   & 267,863   & 6  & 14     & 0  & 471  & 67.6\%  & 97,584  & 10,893 & 33,482 \\
        MAGE-OOD   & 430,630   & 10 & -     & 0  & 267  & 34.86\% & 167,972 & 50,387 & 5,682  \\
        MAGE-OOLLM & 314,817 & - & 7 & 0 & 267  & 31.92\% & 186,636 & 47,988 & 8,022 \\
        \hline
    \end{tabular}
    }
    \caption{Overview of datasets used in the study. Attks is the number of attacks included in the dataset. A.Tokens is the average token length using the Falcon \citenum{almazrouei_falcon_2023} tokenizer. A.Train, A.Val, and A.Test are the average train, validation, test set sizes across all domain splits. The train and validation datasets are class balanced. LLM stats for MAGE-OOD and domain stats for MAGE-OOLLM are not disclosed by the data authors.}
    \label{tbl:dataset}
\end{table*}

The datasets used in this work were used for research purposes. This aligns with their intended use and licenses. The details of the datasets are shown in Table~\ref{tbl:dataset}.

Here we show the mean and standard deviation across three runs, (random seeds 42,43,44) for the methods that are not zero shot or logistic regression based. Note there were three M4GT and four RAID samples where Ghostbuster could not make an inference due to the low number of tokens in the document. For this documents, we infilled a low prediction score indicating human prediction. For the RAID dataset, we used the Binoculars for each document released by \cite{dugan_raid_2024}. 

\begin{table*}
    \centering
    \resizebox{\textwidth}{!}{%
    \begin{tabular}{l|*{8}{c}} 
    &abstracts & books & news & poetry & recipes & reddit & reviews & wiki\\ 
\hline
RoBERTa &93.1$\pm$1.2 & 87.0$\pm$2.1 & 91.4$\pm$3.4* & 95.2$\pm$1.3* & 84.4$\pm$16.9 & 93.9$\pm$1.2* & 90.2$\pm$2.3 & 91.8$\pm$2.8\\ 
Text-Fluor. &71.4$\pm$0.0 & 82.4$\pm$0.0 & 74.9$\pm$0.0 & 70.6$\pm$0.0 & 76.1$\pm$0.0 & 79.2$\pm$0.0 & 73.9$\pm$0.0 & 82.6$\pm$0.0\\ 
UAR-D &71.4$\pm$4.4 & 85.2$\pm$0.8 & 84.5$\pm$1.2 & 73.2$\pm$0.5 & 89.5$\pm$0.8* & 82.4$\pm$0.3 & 84.9$\pm$1.1 & 82.3$\pm$0.2\\ 
UAR-L &89.6$\pm$2.0 & 91.1$\pm$0.2 & 89.8$\pm$0.4 & 76.3$\pm$2.6 & 85.3$\pm$1.2 & 88.8$\pm$0.7 & 88.1$\pm$0.4 & 89.3$\pm$0.5\\ 
\hline
PPL &69.7$\pm$0.0 & 76.8$\pm$0.0 & 69.4$\pm$0.0 & 73.9$\pm$0.0 & 69.6$\pm$0.0 & 76.6$\pm$0.0 & 75.8$\pm$0.0 & 71.3$\pm$0.0\\ 
Binoculars &83.2$\pm$0.0 & 84.3$\pm$0.0 & 80.2$\pm$0.0 & 83.5$\pm$0.0 & 79.4$\pm$0.0 & 83.2$\pm$0.0 & 82.1$\pm$0.0 & 80.2$\pm$0.0\\ 
Fast-DetectGPT &80.0$\pm$0.0 & 80.1$\pm$0.0 & 77.9$\pm$0.0 & 77.1$\pm$0.0 & 75.6$\pm$0.0 & 78.8$\pm$0.0 & 80.0$\pm$0.0 & 79.4$\pm$0.0\\ 
Ghostbuster &88.0$\pm$0.0 & 91.4$\pm$0.0 & 81.6$\pm$0.0 & 88.2$\pm$0.0 & 74.1$\pm$0.0 & 85.0$\pm$0.0 & 81.7$\pm$0.0 & 87.8$\pm$0.0\\ 
\hline
R-SENTRA &94.6$\pm$0.3 & 95.1$\pm$0.3* & 88.4$\pm$0.5 & 92.5$\pm$2.2 & 85.5$\pm$0.9 & 91.7$\pm$0.1 & 87.8$\pm$0.5 & 91.8$\pm$0.3\\ 
SENTRA &95.1$\pm$0.1* & 94.1$\pm$1.6 & 91.3$\pm$0.5 & 95.0$\pm$0.8 & 87.0$\pm$1.5 & 93.7$\pm$0.5 & 90.4$\pm$0.9* & 93.2$\pm$0.7*\\ 
\end{tabular}
    }
    \caption{Mean and standard deviation of the AUROC across random seeds on the RAID dataset.}
    \label{tbl:raid_std}
\end{table*}

\begin{table*}
    \centering
    \resizebox{\textwidth}{!}{%
    \begin{tabular}{l|cccccc} 
&arxiv & outfox & peerread & reddit & wikihow & wikipedia\\ 
\hline
RoBERTa &97.8$\pm$0.3* & 84.9$\pm$2.2 & 82.8$\pm$18.6 & 89.6$\pm$3.9 & 85.5$\pm$2.3 & 88.5$\pm$3.9\\ 
Text-Fluor. &84.7$\pm$0.0 & 84.8$\pm$0.0 & 89.2$\pm$0.0 & 83.9$\pm$0.0 & 78.1$\pm$0.0 & 78.3$\pm$0.0\\ 
UAR-D &73.3$\pm$6.7 & 83.9$\pm$0.2 & 65.7$\pm$1.0 & 86.1$\pm$1.0 & 63.9$\pm$0.6 & 78.9$\pm$2.2\\
UAR-L &93.8$\pm$1.2 & 87.6$\pm$0.6 & 87.1$\pm$0.4 & 80.3$\pm$1.1 & 71.0$\pm$2.4 & 88.4$\pm$0.7\\
\hline
PPL &83.6$\pm$0.0 & 85.7$\pm$0.0 & 94.2$\pm$0.0 & 89.7$\pm$0.0 & 81.7$\pm$0.0 & 87.1$\pm$0.0\\ 
Binoculars &93.1$\pm$0.0 & 82.6$\pm$0.0 & 90.5$\pm$0.0 & 93.8$\pm$0.0 & 79.0$\pm$0.0 & 95.4$\pm$0.0\\ 
Fast-DetectGPT &91.9$\pm$0.0 & 80.3$\pm$0.0 & 88.2$\pm$0.0 & 91.0$\pm$0.0 & 79.1$\pm$0.0 & 93.7$\pm$0.0\\ 
Ghostbuster &94.3$\pm$0.0 & 87.3$\pm$0.0 & 81.9$\pm$0.0 & 95.4$\pm$0.0 & 73.3$\pm$0.0 & 94.5$\pm$0.0\\ 
\hline
R-SENTRA &94.6$\pm$0.5 & 88.4$\pm$0.4* & 94.9$\pm$0.2 & 97.7$\pm$0.3* & 83.9$\pm$1.3 & 97.4$\pm$0.3\\ 
SENTRA &92.3$\pm$1.0 & 88.0$\pm$0.1 & 95.0$\pm$0.3* & 97.7$\pm$0.2 & 87.1$\pm$1.7* & 97.7$\pm$0.3*\\ 
\end{tabular}
    }
    \caption{Mean and standard deviation of the AUROC across random seeds on the M4GT dataset.}
    \label{tbl:m4gt_std}
\end{table*}

\begin{table*}
    \centering
    \resizebox{\textwidth}{!}{%
    \begin{tabular}{l|cccccccccc}
&cmv & eli5 & hswag & roct & sci\_gen & squad & tldr & wp & xsum & yelp\\ 
    \hline
RoBERTa &94.8$\pm$1.0 & 92.9$\pm$0.7 & 87.4$\pm$4.2* & 88.8$\pm$1.0* & 84.3$\pm$6.5 & 93.3$\pm$1.0 & 85.7$\pm$5.1 & 90.3$\pm$1.5 & 74.4$\pm$3.4 & 91.3$\pm$1.6\\ 
Text-Fluoroscopy &62.1$\pm$0.0 & 61.9$\pm$0.0 & 69.5$\pm$0.0 & 71.6$\pm$0.0 & 79.1$\pm$0.0 & 53.3$\pm$0.0 & 73.2$\pm$0.0 & 56.5$\pm$0.0 & 47.8$\pm$0.0 & 64.3$\pm$0.0\\     
UAR-D &80.2$\pm$1.8 & 74.4$\pm$1.7 & 63.5$\pm$2.3 & 61.5$\pm$2.5 & 56.5$\pm$4.7 & 59.6$\pm$3.4 & 60.1$\pm$1.7 & 67.8$\pm$3.3 & 40.5$\pm$0.9 & 70.3$\pm$0.4\\ 
UAR-L &90.1$\pm$0.7 & 81.9$\pm$0.7 & 61.2$\pm$2.4 & 73.5$\pm$1.0 & 80.6$\pm$1.7 & 76.1$\pm$0.8 & 66.3$\pm$2.8 & 88.2$\pm$0.9 & 69.0$\pm$1.9 & 77.5$\pm$1.3\\ 
\hline
PPL &57.9$\pm$0.0 & 61.4$\pm$0.0 & 73.8$\pm$0.0 & 61.2$\pm$0.0 & 49.4$\pm$0.0 & 48.3$\pm$0.0 & 62.9$\pm$0.0 & 59.4$\pm$0.0 & 45.7$\pm$0.0 & 51.9$\pm$0.0\\ 
Binoculars &71.0$\pm$0.0 & 70.2$\pm$0.0 & 59.3$\pm$0.0 & 52.9$\pm$0.0 & 59.7$\pm$0.0 & 55.3$\pm$0.0 & 63.4$\pm$0.0 & 67.2$\pm$0.0 & 57.6$\pm$0.0 & 60.5$\pm$0.0\\ 
Fast-DetectGPT &71.3$\pm$0.0 & 70.1$\pm$0.0 & 66.1$\pm$0.0 & 60.5$\pm$0.0 & 56.4$\pm$0.0 & 57.4$\pm$0.0 & 66.2$\pm$0.0 & 64.5$\pm$0.0 & 54.9$\pm$0.0 & 62.1$\pm$0.0\\ 
Ghostbuster &90.5$\pm$0.0 & 86.0$\pm$0.0 & 66.2$\pm$0.0 & 65.0$\pm$0.0 & 83.6$\pm$0.0 & 78.8$\pm$0.0 & 74.0$\pm$0.0 & 94.1$\pm$0.0 & 72.4$\pm$0.0 & 80.9$\pm$0.0\\ 
\hline
R-SENTRA &98.5$\pm$0.2 & 95.2$\pm$0.7 & 84.6$\pm$0.6 & 87.3$\pm$0.6 & 97.9$\pm$0.1* & 94.1$\pm$0.3* & 93.4$\pm$0.3 & 98.6$\pm$0.3 & 93.8$\pm$1.7 & 94.4$\pm$0.2\\ 
SENTRA &98.6$\pm$0.2* & 95.4$\pm$0.4* & 86.0$\pm$0.3 & 88.2$\pm$0.5 & 97.6$\pm$0.8 & 93.9$\pm$0.6 & 94.1$\pm$0.4* & 98.9$\pm$0.1* & 94.4$\pm$1.0* & 95.1$\pm$0.2*\\ 

\end{tabular}
    }
    \caption{Mean and standard deviation of the AUROC across random seeds on the MAGE-OOD dataset.}
\end{table*}

\begin{table*}
    \centering
    \resizebox{\textwidth}{!}{%
    \begin{tabular}{c|ccccccc} 
&GLM130B & \_7B & bloom\_7b & flan\_t5\_small & gpt.3.5.trubo & gpt\_j & opt\_125m\\ 
\hline 
RoBERTa &77.1$\pm$28.7 & 96.9$\pm$0.6* & 94.6$\pm$1.3* & 69.9$\pm$22.0 & 90.3$\pm$0.5 & 85.4$\pm$19.6 & 95.3$\pm$0.9*\\ 
Text-Fluoroscopy &28.3$\pm$0.0 & 35.7$\pm$0.0 & 42.4$\pm$0.0 & 55.7$\pm$0.0 & 39.0$\pm$0.0 & 41.2$\pm$0.0 & 48.4$\pm$0.0\\ 
UAR-D &80.4$\pm$1.3 & 70.5$\pm$0.6 & 75.3$\pm$0.8 & 66.3$\pm$1.1 & 70.3$\pm$1.8 & 73.3$\pm$0.9 & 65.8$\pm$1.3\\ 
UAR-L &82.8$\pm$0.6 & 71.4$\pm$0.7 & 83.9$\pm$0.5 & 70.7$\pm$0.6 & 77.4$\pm$0.6 & 92.3$\pm$0.2 & 84.4$\pm$1.2\\
\hline
PPL &91.9$\pm$0.0 & 92.8$\pm$0.0 & 41.8$\pm$0.0 & 35.7$\pm$0.0 & 90.5$\pm$0.0 & 25.4$\pm$0.0 & 35.1$\pm$0.0\\ 
Binoculars &94.7$\pm$0.0 & 94.8$\pm$0.0 & 48.1$\pm$0.0 & 52.3$\pm$0.0 & 95.2$\pm$0.0* & 14.7$\pm$0.0 & 32.6$\pm$0.0\\ 
Fast-DetectGPT &3.8$\pm$0.0 & 2.8$\pm$0.0 & 54.5$\pm$0.0 & 53.1$\pm$0.0 & 7.6$\pm$0.0 & 85.8$\pm$0.0 & 57.8$\pm$0.0\\ 
Ghostbuster &88.8$\pm$0.0 & 79.8$\pm$0.0 & 78.1$\pm$0.0 & 54.5$\pm$0.0 & 65.7$\pm$0.0 & 78.1$\pm$0.0 & 34.3$\pm$0.0\\ 
\hline
R-SENTRA &96.8$\pm$0.2 & 93.9$\pm$0.9 & 92.5$\pm$0.8 & 89.9$\pm$0.6 & 93.3$\pm$0.3 & 96.4$\pm$0.3 & 91.5$\pm$1.0\\ 
SENTRA &97.2$\pm$0.3* & 93.3$\pm$1.5 & 94.1$\pm$0.4 & 92.4$\pm$2.0* & 92.6$\pm$1.4 & 97.5$\pm$0.5* & 88.0$\pm$2.3\\ 
\end{tabular}
    }
    \caption{Mean and standard deviation of the AUROC across random seeds on the MAGE-OOLLM dataset.}
\end{table*}

\begin{table*}
    \centering
    \resizebox{\textwidth}{!}{%
    \begin{tabular}{c|cccccccc} 
F1&abstracts & books & news & poetry & recipes & reddit & reviews & wiki\\ 
\hline 
RoBERTa &90.8$\pm$0.4 & 92.7$\pm$1.7 & 94.0$\pm$2.1 & 94.8$\pm$2.1 & 94.2$\pm$2.1 & 94.0$\pm$1.6 & 93.0$\pm$1.9 & 95.5$\pm$0.5\\ 
Text-Fluoroscopy &81.1$\pm$0.0 & 79.6$\pm$0.0 & 83.7$\pm$0.0 & 91.6$\pm$0.0 & 93.9$\pm$0.0 & 73.2$\pm$0.0 & 86.3$\pm$0.0 & 79.2$\pm$0.0\\ 
UAR-D &81.4$\pm$5.7 & 91.0$\pm$0.8 & 89.8$\pm$0.9 & 86.7$\pm$2.7 & 88.7$\pm$1.0 & 89.8$\pm$0.7 & 88.7$\pm$0.8 & 85.7$\pm$0.6\\ 
UAR-L &85.4$\pm$0.6 & 89.4$\pm$0.6 & 88.2$\pm$1.8 & 79.1$\pm$2.4 & 70.5$\pm$2.1 & 89.1$\pm$0.4 & 87.2$\pm$1.0 & 86.8$\pm$0.0\\ 
Ghostbuster &86.5$\pm$0.0 & 87.0$\pm$0.0 & 85.8$\pm$0.0 & 68.7$\pm$0.0 & 84.5$\pm$0.0 & 90.6$\pm$0.0 & 83.8$\pm$0.0 & 78.5$\pm$0.0\\ 
R-SENTRA &90.1$\pm$1.5 & 88.6$\pm$1.3 & 83.2$\pm$3.1 & 87.8$\pm$0.9 & 92.8$\pm$1.7 & 91.5$\pm$3.4 & 88.4$\pm$3.7 & 83.0$\pm$4.1\\ 
SENTRA &88.7$\pm$1.8 & 89.3$\pm$1.3 & 85.4$\pm$2.0 & 88.4$\pm$0.7 & 91.5$\pm$1.0 & 91.9$\pm$0.3 & 90.5$\pm$1.5 & 88.9$\pm$1.6\\ 
\end{tabular}
    }
    \caption{Mean and standard deviation of average F1 on RAID-OOD dataset. A class-1 threshold of 0.50 was chosen for all classifiers.}    
\end{table*}

\begin{table*}
    \centering
    \resizebox{\textwidth}{!}{%
    \begin{tabular}{c|cccccc} 
&arxiv & outfox & peerread & reddit & wikihow & wikipedia\\ 
\hline 
RoBERTa &82.1$\pm$7.5 & 88.5$\pm$1.6 & 88.7$\pm$3.3 & 73.7$\pm$3.2 & 77.6$\pm$0.6 & 56.0$\pm$11.3\\ 
Text-Fluoroscopy &48.4$\pm$0.0 & 84.6$\pm$0.0 & 78.1$\pm$0.0 & 57.2$\pm$0.0 & 68.0$\pm$0.0 & 69.3$\pm$0.0\\ 
UAR-D &52.3$\pm$3.9 & 86.2$\pm$0.3 & 65.7$\pm$1.5 & 78.0$\pm$0.9 & 59.2$\pm$0.6 & 59.5$\pm$1.9\\ 
UAR-L &79.9$\pm$0.5 & 83.4$\pm$0.2 & 83.5$\pm$0.4 & 73.1$\pm$0.9 & 67.0$\pm$1.6 & 78.0$\pm$0.9\\ 
Ghostbuster &86.7$\pm$0.0 & 83.3$\pm$0.0 & 88.6$\pm$0.0 & 87.0$\pm$0.0 & 66.4$\pm$0.0 & 87.9$\pm$0.0\\ 
R-SENTRA &84.7$\pm$0.3 & 83.9$\pm$0.3 & 90.4$\pm$0.2 & 91.5$\pm$0.9 & 75.1$\pm$0.8 & 92.2$\pm$0.3\\ 
SENTRA &82.8$\pm$1.4 & 84.2$\pm$0.2 & 90.4$\pm$0.2 & 89.4$\pm$1.5 & 78.2$\pm$0.3 & 92.0$\pm$0.9\\ 
\end{tabular}
    }
    \caption{Mean and standard deviation of average F1 on M4GT-OOD dataset. A class-1 threshold of 0.50 was chosen for all classifiers.}    
\end{table*}

\begin{table*}
    \centering
    \resizebox{\textwidth}{!}{%
    \begin{tabular}{c|cccccccccc} 
&cmv & eli5 & hswag & roct & sci\_gen & squad & tldr & wp & xsum & yelp\\ 
\hline 
RoBERTa &74.7$\pm$5.3 & 73.7$\pm$3.8 & 66.2$\pm$10.7 & 39.6$\pm$2.8 & 67.0$\pm$4.5 & 59.1$\pm$4.2 & 55.3$\pm$4.5 & 73.6$\pm$6.1 & 47.1$\pm$6.2 & 64.2$\pm$2.7\\ 
Text-Fluoroscopy &55.6$\pm$0.0 & 47.7$\pm$0.0 & 44.8$\pm$0.0 & 66.3$\pm$0.0 & 66.4$\pm$0.0 & 47.5$\pm$0.0 & 39.4$\pm$0.0 & 45.9$\pm$0.0 & 33.7$\pm$0.0 & 50.9$\pm$0.0\\ 
UAR-D &73.2$\pm$1.6 & 67.5$\pm$2.2 & 53.0$\pm$3.0 & 46.9$\pm$3.3 & 45.8$\pm$4.8 & 53.4$\pm$4.8 & 54.4$\pm$2.2 & 57.6$\pm$1.1 & 38.8$\pm$1.3 & 61.1$\pm$0.6\\ 
UAR-L &82.2$\pm$0.7 & 73.9$\pm$0.7 & 46.0$\pm$1.5 & 50.8$\pm$2.9 & 70.4$\pm$2.5 & 63.0$\pm$2.1 & 50.1$\pm$3.5 & 80.4$\pm$1.1 & 62.1$\pm$2.8 & 68.2$\pm$1.3\\ 
Ghostbuster &82.4$\pm$0.0 & 78.7$\pm$0.0 & 60.4$\pm$0.0 & 51.0$\pm$0.0 & 75.8$\pm$0.0 & 70.9$\pm$0.0 & 65.3$\pm$0.0 & 86.2$\pm$0.0 & 65.5$\pm$0.0 & 73.3$\pm$0.0\\ 
R-SENTRA &92.8$\pm$0.7 & 86.8$\pm$1.0 & 76.9$\pm$0.7 & 69.9$\pm$3.1 & 91.5$\pm$0.7 & 85.9$\pm$1.4 & 84.5$\pm$0.4 & 94.1$\pm$0.5 & 86.0$\pm$2.3 & 85.5$\pm$0.9\\ 
SENTRA &92.9$\pm$0.3 & 87.1$\pm$0.6 & 78.5$\pm$0.6 & 69.7$\pm$4.7 & 90.8$\pm$1.1 & 86.0$\pm$0.5 & 84.6$\pm$0.3 & 93.5$\pm$0.9 & 86.5$\pm$1.3 & 86.6$\pm$0.8\\ 
\end{tabular}
    }
    \caption{Mean and standard deviation of average F1 on MAGE-OOD dataset. A class-1 threshold of 0.50 was chosen for all classifiers.}    
\end{table*}

\begin{table*}
    \centering
    \resizebox{\textwidth}{!}{%
    \begin{tabular}{c|ccccccc} 
F1&GLM130B & \_7B & bloom\_7b & flan\_t5\_small & gpt.3.5.trubo & gpt\_j & opt\_125m\\ 
\hline 
RoBERTa &71.3$\pm$22.4 & 88.4$\pm$1.3 & 84.0$\pm$2.7 & 63.1$\pm$19.2 & 80.0$\pm$1.4 & 73.9$\pm$13.6 & 86.6$\pm$1.6\\ 
Text-Fluoroscopy &33.0$\pm$0.0 & 38.5$\pm$0.0 & 36.4$\pm$0.0 & 54.7$\pm$0.0 & 42.1$\pm$0.0 & 40.8$\pm$0.0 & 33.9$\pm$0.0\\ 
UAR-D &71.5$\pm$1.0 & 63.9$\pm$0.5 & 68.2$\pm$0.4 & 61.4$\pm$0.6 & 64.1$\pm$1.7 & 66.6$\pm$0.6 & 60.6$\pm$1.1\\ 
UAR-L &74.9$\pm$1.2 & 63.5$\pm$0.2 & 75.7$\pm$0.5 & 64.4$\pm$0.3 & 69.8$\pm$0.8 & 80.5$\pm$0.3 & 74.7$\pm$1.2\\ 
Ghostbuster &78.8$\pm$0.0 & 71.8$\pm$0.0 & 70.7$\pm$0.0 & 55.4$\pm$0.0 & 60.5$\pm$0.0 & 70.5$\pm$0.0 & 36.0$\pm$0.0\\ 
R-SENTRA &89.7$\pm$0.2 & 86.2$\pm$1.2 & 82.0$\pm$1.8 & 77.7$\pm$3.3 & 83.8$\pm$0.5 & 89.9$\pm$0.2 & 82.5$\pm$1.2\\ 
SENTRA &89.8$\pm$0.5 & 85.1$\pm$1.8 & 84.9$\pm$1.1 & 83.1$\pm$3.0 & 82.6$\pm$1.7 & 91.1$\pm$0.5 & 77.0$\pm$2.9\\ 
\end{tabular}
    }
    \caption{Mean and standard deviation of average F1 on MAGE-OOLLM dataset. A class-1 threshold of 0.50 was chosen for all classifiers.}    
\end{table*}

\section{Computational Complexity} \label{app:complexity}

LLM generators are computationally expensive. Unfortunately, methods that rely on SNTP inputs depend on LLM inference, making it the most costly component of all detection methods studied in this work. However, SENTRA is a relatively small model with only eight Transformer layers, meaning that computational costs at inference are dominated by the production of SNTP inputs. During training, we cache the SNTP sequences so that the LLMs are run only once per sample. SENTRA uses the same LLMs as Binoculars \cite{hans_spotting_2024}, and because the cost of the SENTRA encoder is minimal compared to LLM inference, the overall computational complexity of SENTRA is roughly equivalent to that of the Binoculars method. Refer to Table~\ref{tbl:param_counts} for detailed number of parameters.

Pre-training took approximately 36 hours on a GH200 GPU. We also fine-tuned RoBERTa and SENTRA models on GH200 instances. Fine-tuning for each data split too between .5 and 12 hours.

\begin{table*}[h]
    \centering
    \resizebox{\columnwidth}{!}{%
    \begin{tabular}{l|c}
        Method & Parameter Count \\
        \hline
        RoBERTa-base & 124M \\
        Text Fluoroscopy & 7B (LLM) + 5.1M (FCN) $\approx$ 7B \\
        UAR & 82M \\
        \hline
        Perplexity & 7B (LLM) \\
        Binoculars & 14B (2 LLMs) \\
        Fast-DetectGPT & 2.7B + 6B (2 LLMs) = 8.7B \\
        Ghostbuster & 7B (LLM) + N (LR, N << 7B) $\approx$ 7B \\
        \hline
        SENTRA & 57M (training), 14B (inference) \\
        R-SENTRA & 57M (training), 14B (inference) \\
    \end{tabular}
    }
    \caption{Parameter count of all methods with the actual LLM(s) used in evaluation. LR stands for logistic regression, FCN stands for fully connected network. For Ghostbuster, we observed $N$ to be between 20 to 40.}
    \label{tbl:param_counts}
\end{table*}

\section{Baseline Assumptions and Setups}
\label{sec:appendixA}

This section details the assumptions and setups for all baseline methods if we have made modifications.

For UAR, the original paper compares the distance between the input query and the closest machine support query against the distance between the closest machine support query and the closest human support query. Mathematically speaking, given $Q$ the input query, $H$ the closest human support query, and $\mathbf{M}$ is the seeded machine support queries, the distance $d_{\texttt{Q}} = \min_{m \in \mathbf{M}} [d(Q, m), d(H, m)]$ is used as the prediction. Though this allows $d_Q$ to be directly usable for metric calculation, this is less trivial than a simple nearest neighbor classification where we calculate the percentage of machine support queries among $k$ as the prediction. in our baseline, we employed the simple nearest neighbor approach with $k=10$ and cosine similarity distance measure. For each domain, we randomly sampled 1,000 human and machine texts respectively to form the kNN seed corpus. We did not group texts into episodes and kept episode size of 1 due to the generally longer text lengths compared to twitter posts.

For Text Fluoroscopy, we switched the model from gte-Qwen1.5-7B-instruct to Falcon-7B-Instruct to better align with other baselines by eliminating the effect of model selection. With this change, we modified the input dimension to the feed forward network from 4096 to 4454 due to falcon models hidden state sizes. Despite the possibilities of under-training, we followed their implementation and sampled 160 data points for training, and 20 for validation (during training). The test set metric at the earliest highest validation accuracy was reported.  We also optimized the feature selection script for more efficient batch processing.

For Ghostbuster, we included a minimum accuracy score improvement threshold of $1\mathrm{e}-4$ to avoid over-fitting and allow early stopping for MAGE dataset where we observed significantly more feature selection iterations compared to the other two datasets. In the case of least square convergence failure (\texttt{max\_iter=1000}) in Logistic Regression fitting, the current feature list is taken as the best features for evaluation.

\section{Hyper-parameter Selection}
For RoBERTa, we chose one domain from the MAGE dataset to tune the learning rate. RoBERTa was initialized from RoBERTa base for both the supervised baseline and during contrastive pre-training. With this learning rate, the RoBERTa diverged before the first epoch on one MAGE split and one RAID split. We then turned down the learning rate for these two splits and reran RoBERTa, but the models still diverged. It is possible with additional tuning, RoBERTa could better fit these datasets, but we did not want to pay special attention to the fine-tuning any one method.

For SENTRA, we did a small search over the number of layers, \{2,4,8\}, for the CMV-MAGE data split by looking at the in-domain development loss. We found four layers to work best. We later found SENTRA had trouble fitting the in-distribution validation data of a data. We found that varying the LR and batch size on this dataset had no significant effect, so we kept the defaults of a LR of $1\mathrm{e}-4$ and a batch size of 128 which were the defaults from RoBERTa. We then manually tuned the pre-training model while looking at this in-distribution loss. We ultimately found that eight layers and and two pre-training phases produced the best performance on this in distribution validation dataset.

\end{document}